\documentclass{article}
\usepackage{spconf,amsmath,graphicx}
\usepackage{booktabs}
\usepackage{caption}
\usepackage{subcaption}
\usepackage{booktabs}
\usepackage{graphicx}
\usepackage{comment}
\usepackage{amsmath}
\usepackage{amssymb}
\usepackage{hyperref}
\usepackage{color}
\usepackage{xcolor}

\usepackage{soul}
\newcommand{\R}{\mathbb{R}}
\newcommand{\ra}[1]{\renewcommand{\arraystretch}{#1}}
\title{Is cross-attention preferable to self-attention \\ for multi-modal emotion recognition?}
\name{Vandana Rajan$^{1}$, Alessio Brutti$^{2}$, Andrea Cavallaro$^{1}$}
  \address{$^{1}$Centre for Intelligent Sensing, Queen Mary University of London, UK\\
  $^{2}$Fondazione Bruno Kessler, Trento, Italy }
\begin{document}
\ninept
\maketitle
\begin{abstract}
Humans express their emotions via facial expressions, voice intonation and word choices. To infer the nature of the underlying emotion, recognition models may use a single modality, such as vision, audio, and text, or a combination of modalities. Generally, models that fuse complementary information from multiple modalities outperform their uni-modal counterparts. However, a successful  model that fuses modalities requires components that can effectively aggregate task-relevant information from each modality.
As cross-modal attention is seen as an effective mechanism for multi-modal fusion, in this paper we quantify the gain that such a mechanism brings compared to the corresponding self-attention mechanism. To this end, we implement and compare a cross-attention and a  self-attention model. In addition to attention, each model uses convolutional layers for local feature extraction and recurrent layers for global sequential modelling. We compare the models using different modality combinations for a 7-class emotion  classification task using the IEMOCAP dataset. 
Experimental results indicate that albeit both  models improve upon the state-of-the-art in terms of weighted and unweighted accuracy for tri- and bi-modal configurations, their performance is generally statistically comparable. The code to replicate the experiments is available at \url{https://github.com/smartcameras/SelfCrossAttn}

\end{abstract}

\section{Introduction}
\label{sec:intro}
Emotion recognition (ER) models use one or more modalities, such as audio (language and para-language), images (facial expressions and body gestures) and text (language) to infer the class of underlying emotion~\cite{caridakis2007multimodal}. Multi-modal models are designed to effectively {fuse} relevant information from different modalities and generally outperform uni-modal models~\cite{baltruvsaitis2018multimodal, yoon2020attentive}. ER models may use as input raw signals (speech or face images)~\cite{trigeorgis2016adieu, tzirakis2017end, rajan2019conflictnet} or handcrafted features~\cite{yoon2020attentive, yoon2019speech}. Commonly used speech features are low-level descriptors, such as formants, pitch, log energy, zero-crossing rate and Mel Frequency Cepstral Coefficients (MFCCs)~\cite{yoon2020attentive, yoon2019speech}. Facial expressions can be represented by fixed features based on entities that are always present on the face, such as eyes, mouth and eyebrows and/or transitory features based on temporary entities like wrinkles and bulges~\cite{fasel2003automatic}. Tokenized words can be mapped into linguistic features using word embedding algorithms, such as word2vec~\cite{mikolov} or GloVe~\cite{pennington2014glove}. 

ER models based on Deep Neural Networks (DNNs) may contain convolutional layers to extract local task-relevant components from the input and recurrent layers to facilitate the global sequential modelling  \cite{trigeorgis2016adieu, tzirakis2017end}.
Attention mechanisms~\cite{bahdanau2015neural} integrated in DNN architectures encourage the ER model to focus on task-relevant time instants~\cite{yoon2020attentive,rajan2019conflictnet}. The general purpose of attention mechanism is to provide varying levels of weights to different time-steps in a sequence. There are two types of attention mechanisms, namely self (or intra-modal) attention and cross (or inter-modal) attention. A self-attention mechanism computes the representation of a uni-modal sequence by relating different positions of the same sequence~\cite{rajan2019conflictnet, 8421023}. Cross-modal attention mechanisms use one modality to estimate the relevance of each position in another modality~\cite{tsai2019multimodal}. For example, a self-attention mechanism between 2 recurrent layers can be used to emphasise task-relevant time-steps in an input speech signal~\cite{rajan2019conflictnet}, whereas an iterative multi-hop cross-attention mechanism may select and aggregate information from multi-modal features obtained with Gated Recurrent Unit (GRU) layers~\cite{yoon2020attentive, yoon2019speech}.
Transformers~\cite{vaswani2017attention}, which contain a Multi-Head Attention (MHA) module, are also becoming popular in modelling uni-modal as well as multi-modal emotional data~\cite{rajan2021cross, tsai2019multimodal, sun2021multimodal, huang2020multimodal}. 
Cross-modal transformers use cross-attention to calculate the relevance of each time-step in a target modality representation using a different source-modality~\cite{tsai2019multimodal, huang2020multimodal}. A serial~\cite{tsai2019multimodal, huang2020multimodal} or parallel~\cite{sun2021multimodal} combination of cross and self-attention transformers aims to capture the cross-modal and intra-modal relationships for multi-modal fusion. 
Considering the interest in models combining self and cross attention-based transformer encoders~\cite{tsai2019multimodal, sun2021multimodal, huang2020multimodal}, we conduct the first study comparing the two types of attention mechanisms (without the other transformer components). To understand the differences between the two types of attention mechanisms, we extensively compare a model based only on cross-attention and one based only on self-attention for bi- and tri-modal combinations. 
We compare the two models on the IEMOCAP~\cite{busso2008iemocap} dataset for 7-class emotion classification and conclude that the cross-attention model does not outperform the self-attention model. Nevertheless, both  models improve the state-of-the-art results on tri-modal as well as bi-modal emotion recognition tasks in terms of weighted and unweighted accuracy metrics.

\section{Cross and Self Attention Models}

Self and cross-attention models first process individual modalities using modality-specific encoders. The encoded features are then fed into self or cross Multi-Head Attention (MHA)~\cite{vaswani2017attention} modules, respectively. A global representation of the utterance clip is generated as temporal average at the outputs of each attention module. The resulting features are then concatenated and their mean and standard deviation are obtained using a statistical pooling layer. The concatenation of mean and standard deviation vectors is then fed to fully connected layers. 
The emotional class predictions are obtained through a softmax operation. A detailed explanation is given as follows:

\begin{figure}[t]
    \centering
    \includegraphics[width=7.0cm,height=9.0cm]{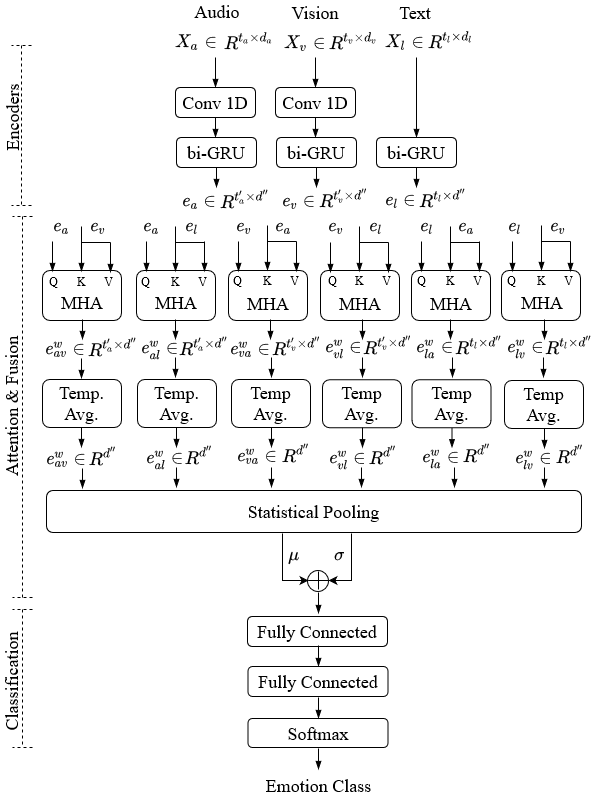}
    \caption{Architecture of the tri-modal cross-attention model.  KEY - MHA: Multi-Head Attention; Temp.: temporal; Avg.: averaging; $\mu$: mean; $\sigma$: standard deviation; $\bigoplus$: concatenation operation.}
    \label{fig:crossarch}
\end{figure}
\begin{figure}[t]
    \centering
    \includegraphics[width=4.5cm,height=5cm]{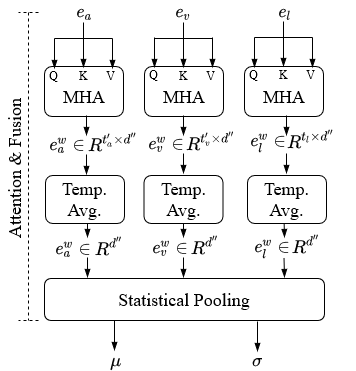}
    \caption{Attention and fusion module in the tri-modal self-attention model. The rest of the model is same as the tri-modal cross-attention model. KEY - MHA: Multi-Head Attention; Temp.: temporal; Avg.: averaging; $\mu$: mean; $\sigma$: standard deviation.} 
    \label{fig:selfarch}
\end{figure}

Let \mbox{$X_a \in \R^{t_a \times d_a}$} be the audio features corresponding to an utterance clip, where $t_a$ is the sequence length and $d_a$ is the feature dimension. The audio encoder consists of a 1D convolution layer followed by a bi-directional GRU. The convolution layer, which refines the input feature sequence by finding task-relevant patterns, operates as follows:
\begin{equation}
    X'_{a}(t') = b(t') + \sum_{k=0}^{t_a-1}(W(t',k) \ast X_a(k)),
\end{equation}
where \mbox{$X_a' \in \R^{t_a' \times d_a'}$} is the output with length $t_a'$ and dimension $d_a'$, \mbox{$t' \in [0, t_a'-1]$}, $\ast$ is the convolution operator, $W$ are the weights and $b$ are the biases associated with the layer. Thus, the convolution layer modifies the sequence length as well as the feature dimension. 

The bi-directional GRU layer models contextual inter-dependence of the features across time. For each element in the sequence, the bi-GRU layer computes the following functions:

\begin{equation} \label{eq:gru}
\noindent
    \begin{cases}
    r_t = \sigma(W_{ir}X'_{a}(t) + b_{ir} + W_{hr}h_{t-1} + b_{hr}), \\
    z_t = \sigma(W_{iz}X'_{a}(t) + b_{iz} + W_{hz}h_{t-1} + b_{hz}), \\
    n_t = \phi_{h}(W_{in}X'_{a}(t) + b_{in} + r_t \odot (W_{hn}h_{t-1} + b_{hn})), \\
    h_t = (1-z_t) \odot n_t + z_t \odot (h_{t-1}),
    \end{cases}
\noindent
\end{equation}
where $h_t$ and $h_{t-1}$ are the hidden states at times $t$ and $t-1$, $X'_{a}(t)$ is the input at time $t$. $r_t$, $z_t$ and $n_t$ are the reset, update and new gates, $W$ and $b$ are the corresponding weights and biases, $\sigma$ and $\phi_{h}$ are the sigmoid and hyperbolic tangent functions and $\odot$ is the Hadamard product. At the output of bi-GRU, the forward and backward hidden states for each time-step are concatenated and the refined audio features can be represented as $e_a \in \R^{t_a' \times d''}$, where $d''$ is twice the number of hidden neurons in the GRU. 

Similar to audio, the vision encoder consists of one 1D convolution layer followed by a bi-GRU layer. If \mbox{$X_v \in \R^{t_v \times d_v}$} represents the vision features corresponding to an utterance, then at the output of vision encoder, the features are refined to \mbox{$e_v \in \R^{t_v' \times d''}$}. For the text modality, the encoder consists of only one bi-GRU layer. The input and output of text encoder can be represented by \mbox{$X_l \in \R^{t_l \times d_l}$} and \mbox{$e_l \in \R^{t_l \times d''}$} respectively.

\begin{table*}[t]
\footnotesize
\centering
\ra{1.0}
\caption{Results of a 7-class emotion classification task presented as mean $\pm$ standard deviation. AMH refers to AMH \cite{yoon2020attentive} for tri-modal models and to MHA \cite{yoon2019speech} for bi-modal models. KEY - A: audio; V: vision; T: text; Self: self-attention model; Cross: cross-attention model. The best results in each row are in bold font. The symbol * refers to the only three results with statistically significant difference between the self and cross models.}
\begin{tabular}{c@{\hskip .1in}|@{\hskip .1in}c@{\hskip .2in}c@{\hskip .2in}c@{\hskip .2in}c@{\hskip .006in}c|@{\hskip .1in}c@{\hskip .2in}c@{\hskip .2in}c@{\hskip .2in}c@{\hskip .006in}c}
        \toprule
        \vspace{2mm}
         & \multicolumn{5}{c|@{\hskip .1in}}{\bf{Weighted Accuracy}} & \multicolumn{5}{c}{\bf{Unweighted Accuracy}} \\
         Modality & MDRE~\cite{yoon2018multimodal} & AMH~\cite{yoon2020attentive, yoon2019speech} & Cross & Self & & MDRE~\cite{yoon2018multimodal} & AMH~\cite{yoon2020attentive, yoon2019speech} & Cross & Self & \\
         \midrule
         T & - & - & - & .474 $\pm$ .030 & & - & - & - & .535 $\pm$ .016 & \\
         V & - & - & - & .454 $\pm$ .019 & & - & - & - & .513 $\pm$ .018 & \\
         A & - & - & - & .365 $\pm$ .018 & & - & - & - & .452 $\pm$ .017 & \\
         \midrule
         T+V & .524 $\pm$ .021 & .526 $\pm$ .024 & \textbf{.567 $\pm$ .022} & .563 $\pm$ .022 & & .579 $\pm$ .015 & .580 $\pm$ .019 & \textbf{.617 $\pm$ .015} & .614 $\pm$ .020 & \\
         T+A & .418 $\pm$ .077 & .491 $\pm$ .028 & .501 $\pm$ .026 & \textbf{.518 $\pm$ .031} & * & .498 $\pm$ .059 & .543 $\pm$ .026 & .562 $\pm$ .017 & \textbf{.574 $\pm$ .018} & *\\
         V+A & .376 $\pm$ .024 & .371 $\pm$ .042 & .481 $\pm$ .024 & \textbf{.483 $\pm$ .026} & & .477 $\pm$ .025 & .471 $\pm$ .047 & .566 $\pm$ .022 & \textbf{.567 $\pm$ .026} & \\
         \midrule
         T+V+A & .490 $\pm$ .056 & .547 $\pm$ .025 & .578 $\pm$ .024 & \textbf{.587 $\pm$ .022} & * & .564 $\pm$ .043 & .617 $\pm$ .016 & .636 $\pm$ .017 & \textbf{.642 $\pm$ .019} & \\
         \bottomrule
\end{tabular}
%
\label{tab:my_label}
\end{table*}

We use the MHA module~\cite{vaswani2017attention} for self and cross-attention modelling. An MHA module consists of multiple such attention operations to capture richer interpretations of the sequence. 
Each MHA module requires 3 inputs, namely, Query ($Q$), Key ($K$) and Value ($V$), each of which is first projected $H$ times to different sub-spaces using linear layers, where $H$ refers to number of heads. Projections for each sub-space $h \in \{0, .... , H-1\}$ can be calculated as

\begin{equation} \label{eq:proj1}
    Q_h = W_h^Q e_m, \\
\end{equation}
\begin{equation} \label{eq:proj2}
    K_h = W_h^K e_m, \\
\end{equation}
\begin{equation} \label{eq:proj3}
    V_h = W_h^V e_m, \\
\end{equation}
where $m \in \{a,v,l\}$ denotes the modality.
In each of these sub-spaces, scaled dot-product attention is performed on the projections. For a sub-space $h$, the attention operation is given as
\begin{equation}
    Att_{h}(Q_{h},K_{h},V_{h}) = \mathtt{Softmax}\left(\frac{Q_{h}K_{h}{^T}}{\sqrt{d_k}}\right)V_{h},
\end{equation}
where $Att_{h}(\cdot)$ and $d_k$ refer to the attention operation in sub-space $h$ and feature dimensionality, respectively. The outputs of all $H$ attentions are concatenated and passed through a linear layer to obtain the final output of an MHA module. 

In the cross-attention model, a source modality is given as $K$ and $V$, whereas a target modality is fed as $Q$ (see Fig. \ref{fig:crossarch}). The intuition behind such an approach is to discover cross-modal interactions by adapting the source modality to the target modality~\cite{tsai2019multimodal}. As an example, let us take the case of audio as target modality and vision as the source modality. The refined audio features \mbox{$e_a \in \R^{t_a' \times d''}$} are transformed to $Q$ using Eq. \ref{eq:proj1} and vision features \mbox{$e_v \in \R^{t_v' \times d''}$} to $K$ and $V$ using Eq.~(\ref{eq:proj2})-(\ref{eq:proj3}). The cross-modal MHA module then maps the vision to the audio modality and outputs vision features adapted to audio \mbox{$e_{av}^{w} \in \R^{t_a' \times d''}$}. Note that the sequence length of the cross-attention weighted output is the same as the target modality audio. With 3 modalities, we have 6 combinations of source-target modalities and hence we use 6 MHA modules. In case of self-attention model, the input sequence corresponding to the same modality is used as $Q$, $K$ and $V$ (see Fig. \ref{fig:selfarch}). 
This helps to capture intra-modal interactions in each modality. For cross-attention model, statistical pooling is done across the concatenation of the temporal averages of 6 cross-modal sequences, whereas for the self-attention model, it is done across the concatenation of the temporal averages of the self-attended sequences of all the 3 modalities.


The classifier for both models is:
\begin{equation} \label{eq:stat}
\hat{y} = \mathtt{Softmax}(f_{\theta_{2}}(f_{\theta_{1}}([\mu \mathbin\Vert \sigma]))), \\
\end{equation}
where $\mu$ and $\sigma$ are the mean and standard deviation obtained from the output of statistical pooling layer, $\mathbin\Vert$ represents concatenation operation, $f_{\theta_{1}}$ and $f_{\theta_{2}}$ denote the 2 fully connected layers with parameters $\theta_{1}$ and $\theta_{2}$, respectively, and $\hat{y}$ denotes the one-hot vector of emotion prediction.

\section{Validation}

In this section, we discuss the dataset and results of using cross- and self-attention models for 7-class bi-modal and tri-modal emotion recognition. We also discuss comparison with state-of-the-art methods and experiments with additional model configurations for both  models.

We use the Interactive Emotional Dyadic Motion Capture (IEMOCAP)~\cite{busso2008iemocap} dataset which contains approximately 12 hours of audio-visual dyadic emotional interactions in acted and spontaneous settings. The dataset, recorded with 5 male and 5 female speakers, includes the ground-truth text transcripts. The labelling of each utterance was determined by majority voting from 3 annotators. There is lack of consensus amongst researchers on the use of IEMOCAP dataset. Some use it for 4 class classification~\cite{tsai2019multimodal} by merging different classes (\textit{happy} and \textit{excited}, \textit{angry} and \textit{frustrated}), while others~\cite{yoon2020attentive, yoon2019speech, yoon2018multimodal, sun2021multimodal} perform 7-class classification. We follow the latter. Since the creators of the dataset did not define a training-testing split, we use the same dataset partition and features as~\cite{yoon2020attentive, yoon2019speech, yoon2018multimodal}. The final dataset contains 7,487 utterances in total  (1,103 \textit{angry}, 1,041 \textit{excited}, 595 \textit{happy}, 1,084 \textit{sad}, 1,849 \textit{frustrated}, 107 \textit{surprise} and 1,708 \textit{neutral}). 
Class sizes smaller than 100 utterances (\textit{fear}, \textit{disgust}, \textit{other}) are eliminated~\cite{yoon2020attentive}. We perform 5-fold cross validation to assess the model performance. Data in each fold are split into training, development, and testing sets (8:0.5:1.5). We train and evaluate the model 10 times (with 10 different random seeds) per fold, and the performance is assessed in terms of weighted accuracy (WA) and unweighted accuracy (UWA) metrics.

For the audio modality, 40D MFCC features (frame size is set to 25 ms at a rate of 10 ms with the Hamming window) are extracted and concatenated with their first and second order derivatives to obtain the final acoustic feature dimension of 120. Audio features are standardised by removing the mean and scaling to unit variance. For vision data, cropped face images of speakers are fed into a ResNet-101~\cite{he2016deep} to obtain 2048D features at a frame rate of 3 Hz. For text modality, 
each word in an utterance is represented by a 300D GloVe~\cite{pennington2014glove} embedding. Note that the modalities are sampled at different rates and the maximum sequence length of audio, vision and text modalities is set to 1,000, 32 and 128 respectively.

The models are implemented using PyTorch~\cite{paszke2017automatic}. The bi-modal and uni-modal versions of the tri-modal models are created by removing components corresponding to the unused modality/modalities. We use Adam~\cite{kingma2014adam} optimiser with a learning rate of 0.001. The learning rate is reduced by a factor 0.1 when the validation loss has stopped decreasing for 10 consecutive epochs. Training is stopped when UWA does not improve in the validation set for 10 consecutive epochs and the model with best validation UWA is used for testing. The batch size is 32 and all models are trained using the categorical cross-entropy loss.

The audio and vision encoders contain one 1D convolution layer each. The kernel size and stride length are both set to 1. The number of input and output channels for audio convolution layer are 1,000 and 500 respectively while for vision they are 32 and 25 respectively. The number of bi-GRU layers for all the 3 modalities is 1. The number of hidden neurons in each bi-GRU layer is 60. The number of attention heads in all MHA modules is 6 and a dropout rate of 0.1 is applied to reduce overfitting. The number of neurons in the first and second fully connected output layers are 60 (same as number of bi-GRU neurons) and 7 (number of output classes) respectively. All parameters were chosen based on the performance on validation set. 

\begin{figure}[t!]
    \centering
    \includegraphics[width=8cm, height=4.8cm]{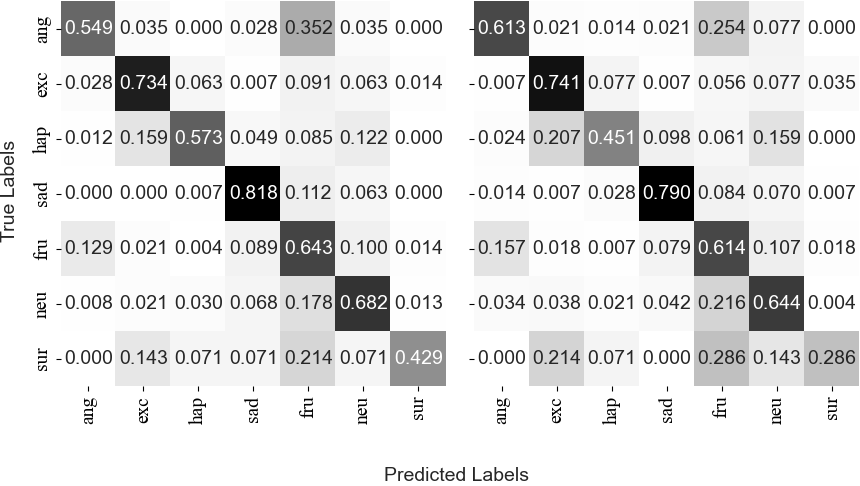}
    \caption{Confusion matrices of self (left) and cross-attention models (right) for tri-modal 7-class classification using a random fold. The emotions classes are abbreviated with their first 3 letters.}
    \label{fig:crossconf}
\end{figure}

Table~\ref{tab:my_label} shows the performance of the self and cross-attention models on 7-class uni-modal, bi-modal and tri-modal emotion recognition tasks. We report the mean and standard deviation obtained across 50 runs (5 folds $\times$ 10 repetitions) for each model. We also applied two-tailed t-test with the null hypothesis that the accuracy values of both self and cross-attention models have identical average (expected) values. 
Comparison of the uni-modal performances shows that the text outperforms the vision and audio modalities. This result is consistent with previous work~\cite{ tsai2019multimodal,yoon2018multimodal}. Since uni-modal performance evaluation is not possible with the cross-modal model, we report results with the uni-modal version of the self-attention model. Among bi-modal models, the combination of vision and text modalities gives the best performance for both models. These results are consistent with previous work~\cite{yoon2019speech, yoon2018multimodal}. Overall, both models provide comparable performances for bi- and tri-modal cases. Self-attention significantly outperforms cross-attention (P value $<$ .05) only for T+A (text and audio) and the WA of T+V+A (text, video, and audio). 

We compare with methods that use the same set of features and dataset partition.
The tri-modal models are compared with AMH~\cite{yoon2020attentive}, the current state-of-the-art model, which uses a combination of uni-modal GRU layers and an iterative attention mechanism\footnote{We use the revised results of AMH, MHA and MDRE from  \url{https://github.com/david-yoon/attentive-modality-hopping-for-SER}. We note that the WA and UWA values were swapped by the authors and we rectify this error in Table~\ref{tab:my_label}.}. Note that the self-attention model exceeds the performance of AMH by 4.0 and 2.5 percentage points (pp) over mean in terms of WA and UWA, respectively. Similar figures for the cross-attention model are 3.1 pp and 1.9 pp. We also compare with MDRE~\cite{yoon2018multimodal}, which uses recurrent layers to model uni-modal signals followed by aggregation and classification using fully connected layers. The better performance of the self and cross-attention models, as well as AMH, compared to MDRE can be attributed to the effectiveness of the attention mechanism. For bi-modal models, we compare with the bi-modal version of AMH called MHA~\cite{yoon2019speech} and MDRE. Again, both  models outperform MHA and MDRE in all the 3 bi-modal cases. Note that we obtain bi-modal results by ablating the tri-modal models and not by fine-tuning for individual bi-modal cases. Also, AMH, MHA and MDRE use prosody features in addition to MFCC features for audio, whereas we use only MFCC features. The state-of-the-art result for text+audio case is obtained by ~\cite{sun2021multimodal} (0.560 WA and 0.612 UWA) which is significantly higher than the bi-modal T+A (text and audio) results. We hypothesize two reasons for this: (1) unlike~\cite{sun2021multimodal}, the bi-modal models are not fine-tuned for the bi-modal cases; (2) \cite{sun2021multimodal} uses transformer encoders that contain additional parameters that might help in learning more complex inter-modal relationships, whereas we use only the multi-head attention mechanism. Nevertheless, both models improve the state-of-the-art tri-modal results of AMH.

Fig.~\ref{fig:crossconf} shows the confusion matrices for the self and cross-attention models.  For both  models we can observe that the classes \textit{angry} and \textit{frustrated} are more often confused with each other, and the class \textit{happy} gets confused with \textit{excited} (these 2 classes are inherently similar). The poor performance of both models on the class \textit{surprise} can be attributed to the fact that this has the smallest sample size in the dataset. These observations are consistent with the previous literature~\cite{yoon2020attentive}.

\begin{table}[t]
\footnotesize
\centering
\ra{1.0}
\caption{Weighted accuracy (WA) and Unweighted accuracy (UWA) for 7-class emotion classification using additional tri-modal model configurations. Self and Cross model results are also shown for comparison. KEY - SP: statistical pooling; Cross-noSP and Self-noSP: cross and self-attention models without SP; Cross+Self:  combination model that concatenates mean and standard deviation vectors from self and cross-attention models.}
\begin{tabular}{l@{\hskip .2in} |@{\hskip .2in} c@{\hskip .2in} |@{\hskip .2in} c}\toprule
        \bf{Model} & \bf{WA} & \bf{UWA} \\
         \midrule
         Cross-noSP & .570 $\pm$ .021 & .634 $\pm$ .015 \\
         Cross & .578 $\pm$ .024 & .636 $\pm$ .012 \\
         Self-noSP & .584 $\pm$ .021 & .638 $\pm$ .019 \\
         Self & \textbf{.587 $\pm$ .022} & \textbf{.642 $\pm$ .019} \\
         Cross+Self & .585 $\pm$ .028 & .642 $\pm$ .020 \\
         \bottomrule
\end{tabular}
\label{tab:sp}
\end{table}

In addition to the two described model configurations, we also experimented with different variations of the tri-modal models. 
We removed the statistical pooling layer from both  models to assess its significance. The outputs from all temporal averaging modules (see Fig.~\ref{fig:crossarch} \& ~\ref{fig:selfarch}) were concatenated and passed to the classifier module. These models are shown as Cross-noSP and Self-noSP in Table~\ref{tab:sp}. We can make two observations. Firstly, the self-attention model outperforms the cross-attention model (P value $<$ .05 for WA) even after ablating statistical pooling. Secondly, the performance of both models decreases without the statistical pooling layer. We also assessed the performance of a combined model created by merging the self and cross-attention models (Cross+Self). The  statistical pooling output from both models were concatenated and fed to a common classifier module. We can see that the performance is similar to that of the self-attention model. This might indicate that the cross-attention model does not contribute any additional, relevant information compared to that of the self-attention model. 

\section{Conclusion}

Intrigued by the popularity of cross-attention mechanism in multi-modal fusion, we compared models based on self-attention  and  on  cross-attention using the IEMOCAP  dataset  for  tri-modal  and  bi-modal  7-class  classification. Results show that there is no meaningful difference between the results of the two  models. Thus, within the context of the dataset and architecture we used, we conclude that cross-attention does not outperform self-attention for multi-modal emotion recognition. Furthermore, both the self and the cross-attention models improve the state-of-the-art in the recognition task. Future work includes investigating the effectiveness of cross and self-attention models for other multi-modal tasks and modalities.

\vspace{.2cm}
\noindent{\bf{Acknowledgement}}. We thank the authors of~\cite{yoon2020attentive, yoon2019speech, yoon2018multimodal} for providing the processed and partitioned IEMOCAP dataset. We acknowledge the use of the ESPRC funded Tier 2 facility, JADE.

\bibliographystyle{IEEEbib}
\bibliography{strings,refs}

\end{document}